\documentclass{article}

\usepackage{PRIMEarxiv}

\usepackage[utf8]{inputenc} 
\usepackage[T1]{fontenc}    

\usepackage{url}            
\usepackage{booktabs}       
\usepackage{amsfonts}       
\usepackage{nicefrac}       
\usepackage{microtype}      
\usepackage{lipsum}
\usepackage{fancyhdr}       
\usepackage{graphicx}       
\graphicspath{{media/}}     

\usepackage{graphicx}
\usepackage{booktabs}
\usepackage{multirow}
\usepackage[accsupp]{axessibility}  
\usepackage{amsmath}   
\usepackage{hyperref}
\usepackage{cleveref}

\title{GlazyBench: A Benchmark for Ceramic Glaze Property Prediction and Image Generation
}

\author{
  Ziyu Zhai, Siyou Li, Juexi Shao, Juntao Yu \\
  Queen Mary University of London \\
  \texttt{\{z.zhai, siyou.li, j.shao, juntao.yu\}@qmul.ac.uk} \\
}

\begin{document}
\maketitle

\begin{abstract}
Developing ceramic glazes is a costly, time-consuming process of trial and error due to complex chemistry, placing a significant burden on independent artists. While recent advances in multimodal AI offer a modern solution, the field lacks the large-scale datasets required to train these models. We propose GlazyBench, the first dataset for AI-assisted glaze design. Comprising 23,148 real glaze formulations, GlazyBench supports two primary tasks: predicting post-firing surface properties, such as color and transparency, from raw materials, and generating accurate visual representations of the glaze based on these properties. We establish comprehensive baselines for property prediction using traditional machine learning and large language models, alongside image generation benchmarks using deep generative and large multimodal models. Our experiments demonstrate promising yet challenging results. GlazyBench pioneers a new research direction in AI-assisted material design, providing a standardized benchmark for systematic evaluation.

\end{abstract}

\section{Introduction}
\label{sec:intro}
The development of ceramic glazes has long relied on empirical paradigms. Complex reactions and phase transitions during high-temperature firing determine the final color, surface texture, and transparency \cite{feng2023phase, wu2021firing}. Consequently, trial and error remains the standard method for achieving functional and desired results. However, this approach is both expensive and time-consuming. It places a significant burden on the development of new glazes, particularly for independent ceramic artists who lack the resources of large manufacturers. Furthermore, the traditional trial-and-error process is highly sensitive to process disturbances \cite{castela2010development, liu2021machine}. This sensitivity results in insufficient reproducibility across different studios and kilns, thereby limiting the systematic exploration of the design space \cite{zhao2023revealing, santos2025temperature}. Machine learning models are increasingly applied in materials science. Despite this progress, the lack of high-quality datasets and standardized evaluation methods remains a serious bottleneck \cite{rumble2017accessing, fujinuma2022big, chakraborty2026survey}. Therefore, building a high-quality dataset and utilizing computational modeling to guide the glaze development process remains an open and challenging question.

\begin{figure}[tb]
  \centering
  \includegraphics[width=\linewidth]{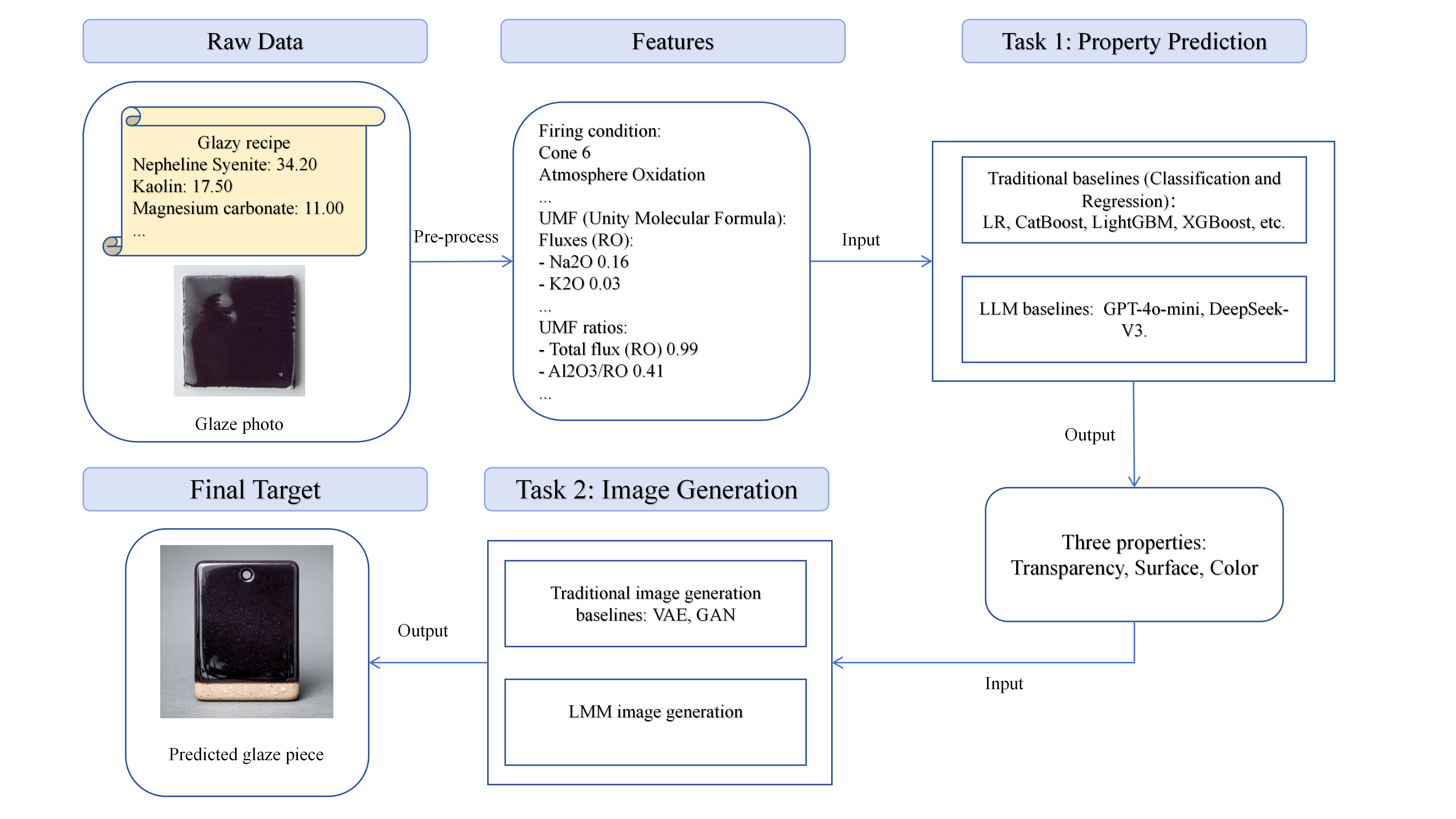}
  \caption{Two-step image generation task}
  \label{fig:intro}
\end{figure}

Researchers have explored interpretable models, such as the Kubelka–Munk (KM) optical model, to successfully predict reflection spectra and CIELAB values for industrial color-matching \cite{bondioli2006color, schabbach2011colouring, schabbach2013color}. In the data-driven domain, neural networks have been used to predict post-firing properties from pigment ratios in industrial wall tiles \cite{romagnoli2008neural}. However, most research focuses on highly specific problems or single glaze systems, such as black, brown \cite{wu2021firing} and bright blue glazes \cite{feng2023phase}, or lustrous layers \cite{imer2016effects}. By relying on fixed raw materials and narrow formula variations, these studies offer strong interpretability but limited transferability when components or firing conditions change. Other work optimizes formulas by adjusting a few components within a fixed kiln type, often treating complex firing dynamics—such as temperature curves and atmospheres—as constants \cite{castela2010development}. Across all these domains, the utilized datasets remain too small or unevenly distributed to effectively train and evaluate modern machine learning models.

In contrast to industry-led research, independent ceramic artists typically rely on books and community resources to explore new glaze recipes. The Glazy platform \cite{glazy} is arguably one of the largest publicly available community resources, making it an ideal source for creating a large-scale glaze design dataset. Like many community-driven platforms, data quality varies significantly between recipes. In this paper, we employ comprehensive data cleaning and standardization methods to transform Glazy's raw data into a high-quality benchmark. We propose a two-step benchmark task that explicitly connects recipe representation, firing context, appearance properties, and image generation. The first step extracts the Unity Molecular Formula (UMF) from raw material information. It combines this formula with the cone rating and firing atmosphere to predict the surface properties of the glaze, including color, surface texture, and transparency. The second step generates a visual representation of the glaze based on these predicted properties. This dual approach enables the model to predict properties across heterogeneous firing contexts while converting performance data into perceptible visual results. Consequently, it supports visual exploration and modification based on specific performance goals. Unlike prior research that primarily focused on single-step composition-to-property tasks, this benchmark emphasizes transferable representations across different kiln environments and provides a complete validation pipeline from performance metrics to final product visuals. The contributions of this paper are organized as follows.

\begin{itemize}
    \item First, Dataset and Standardization. We successfully processed Glazy's recipe and image data by removing duplicates, handling outliers, and ensuring consistent data cleaning. We also unified and standardized the cone ratings, resulting in a reproducible and scalable glaze dataset.

    \item Second, Feature Engineering and Representation. We established a dual representation system spanning from the raw material layer to oxide percentages and Unity Molecular Formula (UMF). Additionally, we constructed physically motivated ratio features, such as SiO$_2$:Al$_2$O$_3$, to enhance the model's interpretability and generalization capabilities.

    \item Third, Multi-task Annotation and Task Setting. We sorted and annotated multi-dimensional properties and appearance labels. These annotations cover color families, continuous color coordinates, surface features, and transparency, thereby supporting multi-task learning scenarios like color regression and surface state classification.

    \item Fourth, High-Quality Image Data and Generation Pipeline. We manually extracted and processed a training set of 4,047 high-quality images based on the standardized dataset, alongside 331 standardized tile sample images. Furthermore, we established a complete image prediction pipeline that translates raw materials and properties into a generated image.
\end{itemize}

\section{Dataset Construction}

\subsection{Data Source and Division}
Data used in this study are obtained from the Glayz website \cite{glazy}. This open-source ceramic glaze database comprises 23,148 real-world glaze recipes. It includes automated color annotations extracted from user-uploaded images. Each recipe provides complete chemical composition information, firing parameters (cone range and atmosphere), glaze surface images, and community-annotated physical properties. The dataset is partitioned into a training set (18,245 samples) and a test set (4,903 samples). The two subsets are strictly disjoint, with no overlap, to prevent data leakage. The authors manually and systematically processed the test set to ensure high data quality.

In addition to the raw chemical composition, we derive augmented features from the original database. First, we use Unity Molecular Formula (UMF) analysis to normalize oxide compositions. This organizes them into flux, stabilizer, and glass-former groups. Second, we include the cone range, which defines the minimum and maximum firing temperatures. Third, we specify the firing atmosphere, denoting oxidation, reduction, or neutral conditions. These features incorporate key ceramic science knowledge. They help capture factors that significantly affect glaze appearance and related properties.

\subsection{Data For Property Prediction}
\cref{tab:coverage_comparison} summarizes the annotation coverage for different prediction targets in the training and test sets. The annotation sources exhibit distinct characteristics. Specifically, transparency and surface labels are collected via enumerated fields filled in by the recipe authors on the website. Transparency is divided into four classes: Opaque, Semi-opaque, Translucent, and Transparent. Surface is divided into nine classes, including Glossy, Matte, and Satin. This structured input promotes label standardization and consistency. In contrast, color annotations are represented by RGB values. The website automatically extracts these values to create color swatches based on photos of fired samples. However, user-uploaded photos are rarely standardized. Consequently, the automatic extraction often yields two prominent colors: the actual glaze color and a background color.

For the training set, we adopt a model-assisted approach to identify the true glaze color. We evaluated four machine learning models on the manually labeled test set to fit the recipe-to-color mapping. 
We selected the two best-performing models, Random Forest and XGBoost. Both models achieved above 94.5\% accuracy after filtering, with a combined voting accuracy of 95.6\% to be used for the final color selection.


The selected Random Forest and XGBoost models were trained on the manually labeled test set. We then used them to predict RGB values for the training samples. We compared the predicted values against the two automatically extracted colors. The color with the smaller prediction distance was selected as the true glaze color. In some cases, both colors yielded similar distances but belonged to different color families (Color family borderline). We treated these samples as unreliable and removed them. The data removal at each stage is summarized in \cref{tab:removal_stats}. This filtering process resulted in 12,175 validated training samples with complete RGB annotations.

\begin{table}[t]
\centering
\caption{Data Removal at Each Stage}
\label{tab:removal_stats}
\begin{tabular}{lrr}
\toprule
\textbf{Filtering Stage} & \textbf{Samples} \\
\midrule
Initial training samples & 18,245 \\

\midrule
\textit{Removed samples (by reason):} & & \\
\quad Missing recipe data & 1,293  \\
\quad Model voting inconsistency & 4,438  \\
\quad Color family borderline & 339  \\
\midrule
Retained samples (passed all filters) & 12,175 \\
\bottomrule
\end{tabular}
\end{table}

Color family categories are deterministically derived from RGB values via color space conversion into nine classes, such as Black and Blue. Consequently, RGB and color family annotations share identical 100\% coverage across all samples. Community-uploaded photos are typically captured under non-standardized lighting conditions. Using a color family classification helps reduce the models' sensitivity to these illumination variations. Automatic extraction often introduces noise, such as background colors, specular highlights, and image artifacts. To avoid this, we manually verified all test set color annotations, including both RGB and color families. This ensures the labels reflect the true fired glaze color. The test set also exhibits a higher annotation density for the transparency and surface tasks. Across all tasks, the category distributions remain consistent between the training and test sets, with a KL divergence below 0.12. This consistency supports adequate representation and reduces the risk of evaluation bias caused by distribution shifts.

\begin{table}[t]
\centering
\caption{Dataset statistics for Property Prediction}
\label{tab:coverage_comparison}
\begin{tabular}{lcccc}
\toprule
\textbf{Feature Type} & \textbf{Train Size} & \textbf{Train \%} & \textbf{Test Size} & \textbf{Test \%} \\
\midrule
\multicolumn{5}{l}{\textit{Prediction Target Features}} \\
\midrule
Transparency (4-class) & 6,584 & 54.1\% & 3,322 & 67.8\% \\
Surface (9-class) & 6,844 & 56.2\% & 3,730 & 76.1\% \\
RGB Color (continuous) & 12,175 & 100.0\% & 4,903 & 100.0\% \\
Color Family (9-class) & 12,175 & 100.0\% & 4,903 & 100.0\% \\
\midrule
\multicolumn{5}{l}{\textit{Input Features (Recipes)}} \\
\midrule
Chemical Composition & 11,384 & 93.5\% & 4,806 & 98.0\% \\
UMF Analysis & 11,474 & 94.2\% & 4,849 & 98.9\% \\
Cone Range (min-max) & 10,747 & 88.3\% & 4,784 & 97.6\% \\
Atmosphere & 10,209 & 83.9\% & 4,729 & 96.5\% \\
\midrule
\textbf{Total Samples} & \textbf{12,175} & - & \textbf{4,903} & - \\
\bottomrule
\end{tabular}
\end{table}

\subsection{Data For Image Generation}
The data used for image generation were manually re-annotated based on the previous test set. This was followed by systematic quality control and screening. Guided by the outcomes of this second round of annotation, we selected the highest-quality samples to constitute the image test set. Images satisfying the baseline inclusion criteria from the first round of annotation formed the image training set. Ultimately, the image training set contains 4,490 samples, and the test set contains 443 samples. Each sample consists of an original image and several key attribute fields.

\begin{figure}[tb]
  \centering
  \includegraphics[width=\linewidth]{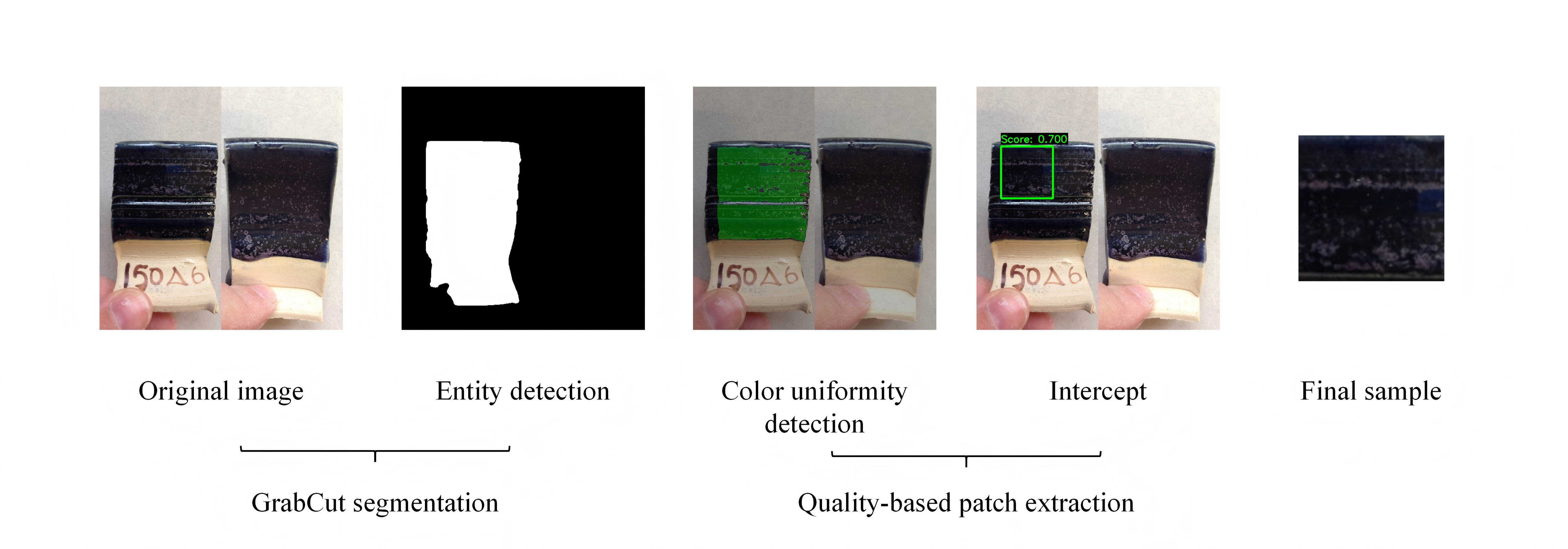}
  \caption{Image region extraction pipeline.}
  \label{fig:image_pre_process}
\end{figure}

The pre-processing pipeline includes integrity verification, region extraction, normalization, and dataset partitioning. First, we removed samples with missing image files or incomplete key attribute fields. All retained samples were required to contain these necessary fields. This strict requirement prevents missing supervision signals from adversely affecting the learning objective. Next, we extracted local regions representing the visual characteristics of the glaze surface from the original images, as shown in \cref{fig:image_pre_process}. These regions were uniformly resized to a fixed input resolution. This standardization reduces the distribution shift induced by variations in composition and scale during image capture. In parallel, we performed image-level validation by checking pixel value ranges and size validity. This step excluded failures caused by corrupted files or abnormal encoding.

\begin{table}[t]
\centering
\caption{Image data automated filtering statistics}
\label{tab:image_filtering_stats}
\begin{tabular}{lrrrr}
\toprule
\multirow{2}{*}{\textbf{Stage}} & \multicolumn{2}{c}{\textbf{Training Set}} & \multicolumn{2}{c}{\textbf{Test Set}} \\
\cmidrule(lr){2-3} \cmidrule(lr){4-5}
& \textbf{Samples} & \textbf{Retention} & \textbf{Samples} & \textbf{Retention} \\
\midrule
Raw samples & 4,490 & 100.0\% & 443 & 100.0\% \\
After GrabCut segmentation & 4,047 & 90.13\% & 438 & 98.87\% \\
After quality patch extraction & 2,323 & 51.74\% & 328 & 74.04\% \\
\bottomrule
\end{tabular}
\end{table} 

Finally, we used GrabCut segmentation \cite{rother2004grabcut, fu2025digital, wei2020experimental} and a Quality-based patch extraction algorithm (LBP Texture Analysis \cite{ojala2002multiresolution}, Sobel Edge Detection \cite{gao2010improved}, etc.) to process this pipeline. \cref{tab:image_filtering_stats} summarizes the filtering statistics. Starting from 4,490 raw training samples and 443 test samples, GrabCut achieved a 90.13\% segmentation success rate for the training data and 98.87\% for the test data. Quality-based patch extraction further retained 2,323 training samples and 328 test samples. The higher retention rate in the test set reflects the superior image quality of the manually curated data. The primary reasons for data removal included segmentation failures, insufficient quality scores, incomplete metadata fields, and inadequate fill ratios. Specifically, segmentation failed for 443 training samples and 5 test samples. The final filtered dataset achieves strong quality metrics. Overall, 83.6\% of the samples scored above 0.6. The average quality scores were 0.688 for the training set and 0.712 for the test set.

\section{Task and Evaluation Settings}
Directly generating glaze images from ingredient lists or a unified molecular formula is highly challenging. Therefore, we decompose the problem into two sequential tasks. First, we predict key glaze properties, including transparency, surface finish, and color. Second, we generate glaze images conditioned on these predicted or given properties.

\subsection{Property Prediction Task Settings}
The dataset supports four core prediction tasks that cover the major visual and physical properties of ceramic glazes. The transparency task employs a four-class classification scheme. This scheme reflects the optical transmission characteristics of ceramic glazes. The Opaque family (Opaque and Semi-opaque) accounts for 70.4\% of the training samples. Whereas Transparent and Translucent categories account for 16.3\% and 13.3\%, respectively (see \cref{tab:transparency_dist} in Appendix B for details). The category distributions between the training and test sets remain consistent.

The surface task contains nine fine-grained labels.  These labels reflect different surface finishes, ranging from highly reflective to completely matte. This task suffers from severe class imbalance. The Glossy class dominates with 49.1\% of the samples, while Stony Matte accounts for only 1.8\% (see \cref{tab:surface_dist} in Appendix B for details). Despite this imbalance, we retain the original nine-class scheme in the benchmark. This preserves fine-grained distinctions that are meaningful to ceramic artists. Future baseline experiments may explore class-merging strategies to mitigate this imbalance.

The color family task categorizes glaze colors into nine semantic categories based on their dominant hue. The distribution is moderately imbalanced. Orange is the most common at 36\% in training, and Purple is the rarest at 0.6\% (see \cref{tab:color_family_dist} in Appendix B for details). Color family labels are directly derived from RGB values through color space conversion, achieving identical 100\% coverage. The training set color family assignments rely on validated RGB values. Meanwhile, the test set labels are completely manually verified to ensure accurate color classification.

RGB color regression is a continuous multi-output regression task. It predicts the three-channel RGB values of the glaze's dominant color within the range of [0, 255]. Both the training and test sets achieve 100\% RGB coverage. All values are validated through either model-based selection for the training set or manual verification for the test set.

\cref{tab:tasks_overview} summarizes the four prediction tasks in the benchmark dataset. All tasks predict the apparent properties of glazes after firing based on recipe information, which includes ingredient composition and the normalized chemical composition of 47 oxides. The effective sample sizes for each task vary under the fixed train-test split due to differences in annotation coverage. Transparency (Task A) and surface (Task B) have a coverage of around 58\% to 61\%. In contrast, the color-related tasks (C1, C2) achieve complete coverage at 100\% because samples lacking reliable color annotations were excluded from the dataset. This complete color coverage enables a robust evaluation of color prediction models. The lower coverage for transparency and surface reflects the inherent difficulty of obtaining consistent annotations for these subtle material attributes.

\begin{table}[tb]
  \caption{Glayz benchmark task configuration overview}
  \label{tab:tasks_overview}
  \centering
  \setlength{\tabcolsep}{3pt}
  \renewcommand{\arraystretch}{1.1}
  \small
  \begin{tabular}{@{}l l c l r r c@{}}
    \toprule
    \textbf{ID} & \textbf{Task} & \textbf{Type} & \textbf{Label Space} & \textbf{Train} & \textbf{Test} & \textbf{Coverage} \\
    \midrule
    A  & Transparency & Classification & 4 classes               & 6{,}584  & 3{,}322  & 58.0\% \\
    B  & Surface      & Classification & 9 classes               & 6{,}844  & 3{,}730  & 61.9\% \\
    C1 & RGB Color    & Regression     & \((r,g,b)\in[0,255]^3\) & 12{,}175 & 4{,}903 & 100.0\% \\
    C2 & Color Family & Classification & 9 classes               & 12{,}175 & 4{,}903 & 100.0\% \\
    \bottomrule
  \end{tabular}
\end{table}

\subsection{Image Generation Task Settings}
The core task of this study is image generation. This task synthesizes glaze images conditioned on surface type, transparency, and target RGB color. We formulate this as learning the conditional distribution $p(x\mid c)$, where $x$ denotes the generated image and $c$ denotes the condition vector. However, because the first task is highly challenging, our image generation task directly uses real properties for training. The model receives surface, transparency, and RGB values as input conditions to generate realistic 128x128 glaze appearance images. This generation task represents our ultimate goal: enabling virtual glaze design and appearance previews without conducting actual firing experiments.
The generation task utilizes a vector that combines categorical properties (surface type, transparency, color family) with continuous RGB values. The surface condition spans nine fine-grained finish types, such as Glossy, Semi-glossy, and Satin. The transparency condition covers four optical states, and the RGB triplet specifies precise color coordinates.

\begin{table}[tb]
  \caption{Image generation dataset statistics. }
  \label{tab:image_generation_overview}
  \centering
  \begin{tabular}{@{}lrrr@{}}
    \toprule
    \textbf{Split} & \textbf{Total} & \textbf{Effective }& \textbf{ Image Resolution}  \\
    \midrule
    Training & 4{,}490 & 2{,}323 & 128×128 RGB \\
    Test & 443 & 328 & 128×128 RGB  \\
    \bottomrule
  \end{tabular}
\end{table}

\cref{tab:image_generation_overview} shows the final number of samples containing high-quality images. The dataset creation follows a two-stage annotation strategy to ensure quality. The first-round annotations, consisting of 4,490 samples, prioritize coverage to capture diverse glaze types from validated recipes. The second-round quality selection, consisting of 443 samples, applies strict visual quality criteria. This step retains only the samples where the selected color patch accurately represents the overall glaze appearance. As shown in \cref{tab:image_generation_overview}, complete attribute coverage for surface, transparency, and color is achieved for 2,323 training samples and 328 test samples. Samples with missing attributes are excluded from model training to ensure consistent conditioning, although they remain available for semi-supervised extension experiments. The "effective" column indicates the final volume of data utilized in our experiments.


\subsection{Evaluation Metrics}
This benchmark uses a predefined, fixed train-test split without random re-partitioning or cross-validation. This ensures evaluation stability and reduces the risk of data leakage. For each task, models are trained on the training set and evaluated on the corresponding annotated subset of the test set. The evaluation pipeline enforces basic validity checks. These include consistency between predicted labels and the task taxonomy, RGB range constraints of [0, 255], and strict mutual exclusivity of train and test IDs.

The benchmark comprises two complementary tasks to comprehensively evaluate glaze property modeling. The first is property prediction from recipes, which predicts visual properties given a chemical composition. The second is image generation, which synthesizes realistic glaze images conditioned on target attributes.

\textbf{Task 1: Property Prediction}. The benchmark includes four prediction tasks: three classification tasks (transparency, surface, and color family) and one regression task (RGB color). For the classification tasks, we use micro F1 and accuracy to assess overall correctness and class-balanced performance. For the RGB color regression, mean absolute error (MAE) was used.


\textbf{Task 2: Image Generation.} The evaluation framework employs a multi-dimensional system divided into sample-level and distribution-level metrics to comprehensively assess generative model performance. Sample-level metrics measure individual image quality. This includes perceptual similarity, calculated via the LPIPS distance to real samples, and color consistency, calculated via the RGB Euclidean distance $d_{\mathrm{RGB}}=\|\overline{\mathrm{RGB}}(\hat{x})-\mathrm{RGB}_{\mathrm{target}}\|_2$ between the generated and target colors. Distribution-level metrics evaluate overall generation quality. This includes the Fr\'echet Inception Distance (FID) to measure distribution discrepancy in a pretrained Inception feature space. It also includes generation diversity, measured by the average pairwise LPIPS under the same condition, to assess the model's ability to avoid mode collapse.


\section{Baseline Methods and Results}

\subsection{Property Prediction}
We established baseline performance on GlazyBench using two approaches. The first approach utilizes supervised learning with traditional machine learning models. The second approach employs zero-shot and few-shot prediction with Large Language Models (LLMs). Traditional baselines learn from labeled training data, whereas LLMs make predictions directly on the test set.

\begin{table}[tb]
\centering
\caption{Traditional baseline performance}
\label{tab:traditional-baselines}
\small
\begin{tabular}{llcc}
\toprule
\textbf{Task} & \textbf{Model} & \textbf{Accuracy} & \textbf{Micro F1} \\
\midrule
\multirow{6}{*}{\parbox{2.5cm}{Transparency\\(4-class)}} 
& RandomForest & 0.434 & 0.262 \\
& LogisticRegression & 0.426 & 0.261 \\
& CatBoost & \textbf{0.525} & \textbf{0.530}  \\
& LightGBM & 0.431 & 0.262  \\
& XGBoost & 0.432 & 0.262 \\
\midrule
\multirow{5}{*}{\parbox{2.5cm}{Surface\\(9-class)}} 
& RandomForest & 0.180 & 0.152  \\
& LogisticRegression & 0.016 & 0.008 \\
& CatBoost & \textbf{0.421} & \textbf{0.444} \\
& LightGBM & 0.097 & 0.073  \\
& XGBoost & 0.108 & 0.071 \\
\midrule
\multirow{5}{*}{\parbox{2.5cm}{Color Family\\(9-class)}} 
& RandomForest & \textbf{0.270} & 0.119 \\
& LogisticRegression & 0.225 & 0.138 \\
& CatBoost & 0.255 & 0.142  \\
& LightGBM & 0.263 & 0.129  \\
& XGBoost & 0.258 & \textbf{0.151}  \\
\midrule
&&\textbf{MAE}& \\
\midrule
\multirow{4}{*}{\parbox{2.5cm}{RGB Color\\(regression)}} 
& RandomForest & 40.30& - \\
& CatBoost  & \textbf{42.20}& - \\
& LightGBM &  40.14& -  \\
& XGBoost &  40.88 & - \\
\bottomrule
\end{tabular}
\end{table}

\subsubsection{Traditional Machine Learning Baselines and Results}
All traditional baselines use the Unity Molecular Formula (UMF) combined with firing parameters as input. The UMF normalizes the glaze composition, while the firing parameters include the cone range and atmosphere.

Due to the scarcity of glaze datasets and related studies, we selected our baseline models by referring to similar research. Specifically, we examined property prediction studies for silicate compounds, such as glass and volcanic rocks. Ultimately, we selected five models commonly used in this domain as our baselines: Logistic Regression (LR) \cite{ueki2018geochemical}, Random Forest (RF) \cite{zhang2020data,ahmmad2021artificial,trott2022random, mues2025using}, XGBoost \cite{ruiyi2021lithology, xie2025prediction}, LightGBM \cite{feng2025application, zhang2025geochemical}, and CatBoost \cite{belciu2025ensemble, vasic2025advanced, bo2022real}. LR provides linear and kernel baselines with interpretable decision boundaries. RF, XGBoost, LightGBM, and CatBoost represent tree ensemble methods with varying optimization strategies. All models use default scikit-learn or library-specific hyperparameters without task-specific tuning, ensuring reproducibility and fair comparison.

\cref{tab:traditional-baselines} summarizes the performance across the four prediction tasks. CatBoost achieves the highest accuracy on transparency (52.5\%) and surface (42.1\%) classification, outperforming the linear baselines by substantial margins. This gap indicates the presence of nonlinear composition-property relationships that tree ensembles can capture, but linear models cannot.

Color family classification proves challenging across all models. The best accuracy is 27.0\% (RF) and the best F1 score is 0.15 (XGBoost). For RGB regression, Random Forest and CatBoost perform similarly (MAE 42.20). This suggests diminishing returns from gradient boosting on this continuous prediction task. Interestingly, the models' performance in RGB regression appears better than in color family classification. However, this merely reflects that RGB distance alone is insufficient for accurately predicting discrete color categories.

\begin{table}[tb]
  \caption{LLM performance comparison across zero-shot and few-shot settings.}
  \label{tab:llm-comparison}
  \centering
  \begin{tabular}{@{}llcccc@{}}
    \toprule
      \textbf{Task}&\textbf{Model} & \multicolumn{2}{c}{\textbf{Zero-shot}} & \multicolumn{2}{c}{\textbf{Few-shot (5-shot)}} \\
     &  &  \textbf{Accuracy}&\textbf{Micro-F1} & \textbf{Accuracy}&\textbf{Micro-F1} \\
    \midrule
    \multirow{3}{*}{Transparency} & GPT-4o-mini & 0.343 &  0.324 & 0.394 & 0.382 \\
    &DeepSeek-v3& 0.326 &  0.312 & \textbf{0.413} &  \textbf{0.409} \\
    &Claude-Sonnet-4.5& \textbf{0.400} &  \textbf{0.382} & 0.380  & 0.395 \\
    \midrule

    \multirow{3}{*}{Surface} & GPT-4o-mini  & 0.270 &  0.224 & 0.402 &  0.373 \\
    &DeepSeek-v3& 0.195 &  0.180 & 0.407 &  \textbf{0.392} \\
    &Claude-Sonnet-4.5& \textbf{0.330} &  \textbf{0.325} & \textbf{0.486}  & 0.379 \\\midrule

    \multirow{3}{*}{Color Family}  &GPT-4o-mini  & \textbf{0.199} &  \textbf{0.176} & 0.169 &  \textbf{0.141} \\
    &DeepSeek-v3 & 0.174 &  0.137 & \textbf{0.175} &  0.138 \\
    &Claude-Sonnet-4.5& 0.179 & 0.139 & 0.155 & 0.121 \\
    
  \bottomrule
  \end{tabular}
\end{table}

\subsubsection{Large Language Model Baselines and Results}
To assess the potential of Large Language Models for glaze property prediction, we tested three LLMs on transparency, surface, and color family classification using zero-shot and few-shot prompting. We strictly restricted the model outputs, requiring them to return the predicted results directly. Because the most advanced models include complex reasoning processes that require intricate prompt design, we selected GPT-4o-mini, DeepSeek-V3.2, and Claude Sonnet 4.5. These models can directly return the required prediction values. The recipe information is provided as text input, containing the chemical composition, UMF ratios, and firing conditions. 

\cref{tab:llm-comparison} summarizes the results. In the zero-shot setting, the three models span a range of 32--40\% accuracy on transparency. Claude Sonnet 4.5 achieves the highest zero-shot transparency accuracy (40.0\%), followed by GPT-4o-mini (34.3\%) and DeepSeek-V3.2 (32.6\%). Surface accuracy is lower across the board (19--33\%), and color family remains the hardest task (17--20\%). Few-shot prompting with five examples yields notable gains, particularly on surface classification. Here, Claude Sonnet 4.5 reaches 48.6\%, surpassing both GPT-4o-mini (40.2\%) and DeepSeek-V3.2 (40.7\%). DeepSeek-V3.2 shows the strongest few-shot gain on transparency, while Claude's performance actually drops in the few-shot setting. Color family classification remains highly challenging for all three models, even with few-shot learning.

These results demonstrate that modern LLMs can achieve performance comparable to simple linear baselines through in-context learning alone, without any task-specific training. Model rankings shift between tasks and settings. However, all the evaluated LLMs failed to achieve highly competitive results. This indicates that domain-specific fine-tuning or specialized feature engineering is necessary for strong performance on materials science prediction tasks. Detailed prompt designs and implementation specifications are provided in Appendix C.

\begin{table}[tbp]
  \centering
  \caption{Color consistency performance of two types of baseline models on the test set (color distance, the lower the better)}
  \label{tab:model_comparison}
  \begin{tabular}{lcccc}
    \toprule
    \textbf{Model} & \textbf{Mean} & \textbf{SD} & \textbf{Median} & \textbf{Pass Rate} \\
    \midrule
    Conditional VAE  & 134.49 & 102.27 & 112.89 & 46.6\% \\
    Lightweight GAN  & \textbf{72.31} & \textbf{64.92} & \textbf{46.54} & \textbf{75.9\%} \\
    \bottomrule
  \end{tabular}
\end{table}

\subsection{Image Generation Baseline Methods and Results}
\subsubsection{Traditional Machine Learning Baselines and Results}
We implemented two representative baseline methods for comparative evaluation.

First is conditional variational autoencoder (cVAE), which introduces a latent variable to model image variability. The encoder learns a conditional posterior over latent codes from the image and its condition, and the decoder reconstructs the image from sampled codes. Training maximizes the evidence lower bound (ELBO) \cite{yan2016attribute2image,yamagiwa2025analytical,wang2025lightweight}. cVAE-based approaches have been applied to glass damage detection \cite{riedel2022automated} and quantitative studies of cement paradigm transformation \cite{mao2025quantitative}.

Another is lightweight GAN based on WGAN-GP. The generator produces images from noise conditioned on target attributes, while the discriminator assesses the realism of image–condition pairs. The Wasserstein objective with a gradient penalty enforces Lipschitz continuity and stabilizes training \cite{ding2021ccgan,jin2019image,li2020sar}. Related applications include glaze-wear analysis \cite{zhang2021understanding} and stylistic glaze image generation \cite{chen2020cantonese}.

Both models use convolutional architectures with approximately 128×128 resolution outputs. The condition vector encodes surface type, transparency, target RGB color, and firing atmosphere as a 25-dimensional feature. Detailed architecture specifications, mathematical formulations, hyperparameter configurations, and training procedures are provided in Appendix D.

In terms of color consistency, the Lightweight GAN markedly outperforms the conditional VAE on the test set (\cref{tab:model_comparison}). The mean color distance decreases from 134.49 to 72.31, the median from 112.89 to 46.54, and the standard deviation from 102.27 to 64.92, indicating both improved accuracy and reduced variability. Using a color-distance threshold of 100, the GAN achieves a 75.9\% excellent rate versus 46.6\% for the VAE. But despite these quantitative gains, visual inspection shows that generated images remain low quality and fall short of the fidelity required for production-level tile prediction.

\begin{figure}[tb]
  \centering
  \includegraphics[width=\linewidth]{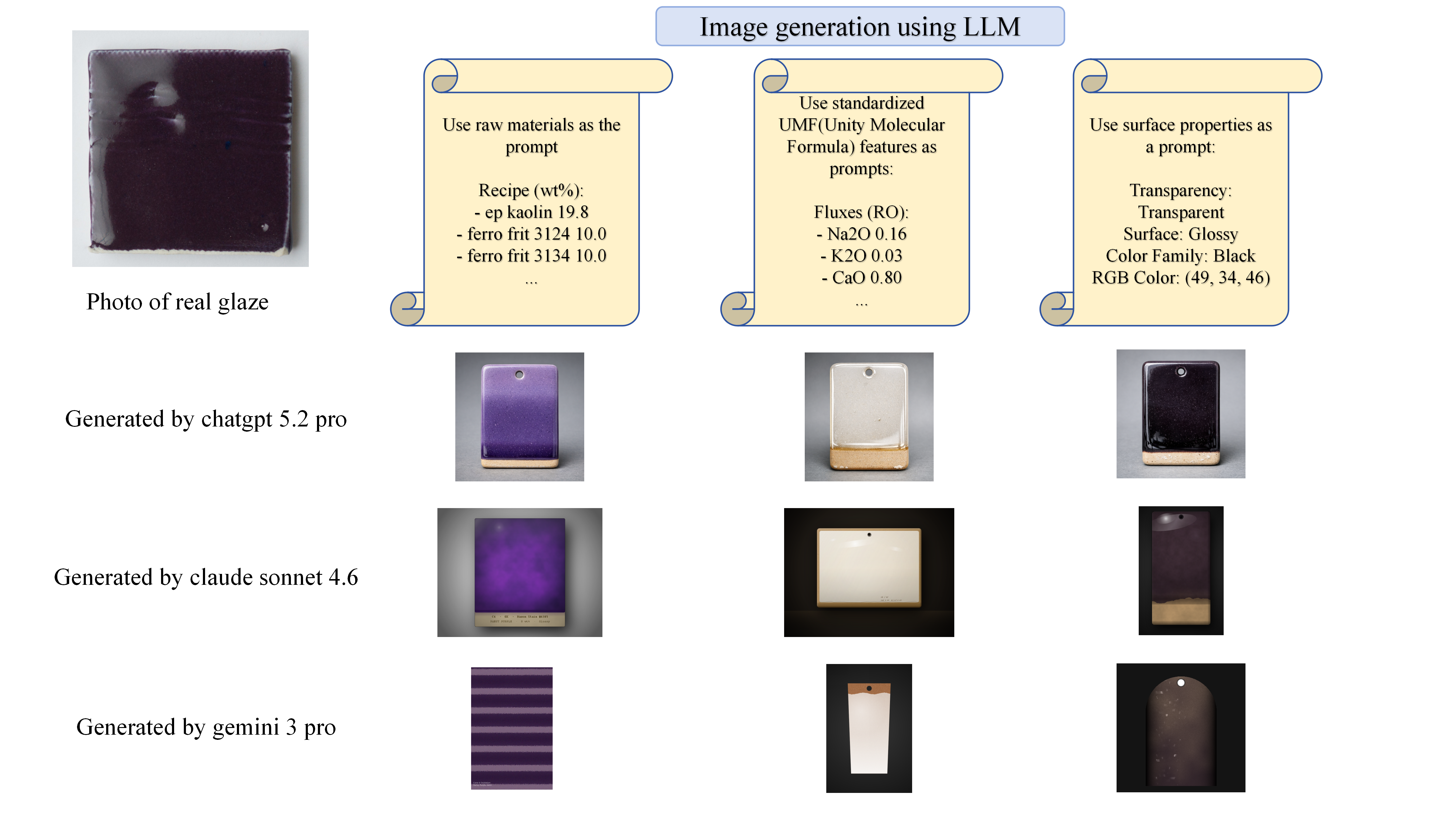}
  \caption{LLM's image generation results under three different prompt conditions}
  \label{fig:LLM_gen_image}
\end{figure}

\subsubsection{Large Multimodal Model Baselines}
As shown in \cref{fig:LLM_gen_image}, we used three types of prompts as conditional inputs for Large Multimodal Models (LMMs): (1) Raw material information (wt\%), (2) Normalized UMF features, and (3) Surface attributes (transparency, gloss, and RGB color). We evaluated the latest versions of three current LMMs for comparison. A very obvious phenomenon emerged during testing. When the input consists of the raw material formula or UMF, the outputs generally fall into a coarse-grained color and style space. There is a significant deviation from real glazed tiles regarding color tone, transparency, and gloss reflection characteristics. It is highly difficult for the models to stably infer the final appearance from chemical composition alone. 

Conversely, when the surface attributes are provided directly, the generated results are significantly closer to the real samples in both dominant color and highlight texture. This indicates that current general-purpose models excel at rendering generation based on high-level visual semantics and explicit appearance constraints. However, they struggle with predictive generation based on underlying material mechanisms. This phenomenon reveals that current LMMs lack a reliable, causal understanding of the "recipe-to-appearance" process. The key reason for this shortfall is the scarcity of training data that strictly pairs professional variables with real ceramic tile images. Therefore, constructing datasets that rigorously align recipes, UMFs, process conditions, standardized imaging, and quantifiable appearance labels is a critical foundation. Such datasets are essential for moving models away from prior-driven schematic generation and toward verifiable, prediction-based generation grounded in material science.

\section{Conclusion and Future Work}
We have proposed the first benchmark for glaze design, named GlazyBench. This benchmark collates 23,148 real glaze formulations. It covers two core tasks: property prediction and image generation. Among these, 4,903 test samples were strictly selected by human annotators to maintain a balanced distribution in the train/test split. This careful curation supports repeatable and comparable system evaluations. Based on this dataset, we constructed a two-step prediction pipeline and systematically compared two representative methods: traditional machine learning models and large language/multimodel models (LLMs/LMMs). For the property prediction task, we implemented and evaluated traditional baselines, including Random Forest and XGBoost, alongside LLMs. For the image generation task, we provided deep generative model baselines and large multimodal model (LMM) example schemes. These comparisons effectively illustrate the capability boundaries of different AI paradigms. 

Experimental results show that general multimodal models suffer from significant conditional interpretability collapse during condition mapping. Consequently, their outputs are rarely stable or verifiably constrained by the formula conditions. Simultaneously, we verified the feasibility of predicting glaze properties directly from formula components. We also revealed key limitations in existing methods regarding data representation, feature engineering, and model architectures. These findings emphasize the importance of constructing standardized datasets. Such datasets must include formulas, Unity Molecular Formulas (UMFs), firing conditions, and quantifiable appearance labels that strictly match the corresponding glaze images. We hope that GlazyBench will drive this field from suggestive generation toward verifiable prediction and generation. It provides a unified and reliable foundation for future model design, evaluation protocols, and interpretability research.

Despite these contributions, the current release has several limitations, primarily concerning data coverage and generalization. Although GlazyBench covers a large number of real recipes and process records, the data source inevitably influences the recipes, firing conditions, and collection methods. This influence may introduce biases toward specific geographical regions or material systems. Therefore, the models' generalization capabilities across different kilns, recipe families, and firing conditions require more systematic verification. This is especially true for novel raw material systems and extreme firing parameters. 

Additionally, noise exists within the appearance labels and image observations. Labels indicating transparency, surface type, and color range are frequently affected by measurement errors, lighting variations, camera differences, and inconsistent naming standards. These issues are particularly prevalent in community-sourced data. Currently, there is insufficient rigorous verification to guarantee the absolute precision of the data, which inherently caps the upper limit of achievable model performance. We aim to address this issue in the next release of the dataset by incorporating more standardized sources, such as published books and professional material archives.

%
%
\bibliographystyle{splncs04}
\bibliography{references}

@String(TOG   = {ACM Trans. Graph.})

@String(TOG   = {ACM TOG})

@article{bondioli2006color,
  title={Color matching algorithms in ceramic tile production},
  author={Bondioli, Federica and Manfredini, Tiziano and Romagnoli, Marcello},
  journal={Journal of the european ceramic society},
  volume={26},
  number={3},
  pages={311--316},
  year={2006},
  publisher={Elsevier}
}

@article{schabbach2011colouring,
  title={Colouring of opaque ceramic glaze with zircon pigments: Formulation with simplified Kubelka--Munk model},
  author={Schabbach, LM and Bondioli, Federica and Fredel, MC},
  journal={Journal of the European Ceramic Society},
  volume={31},
  number={5},
  pages={659--664},
  year={2011},
  publisher={Elsevier}
}

@article{schabbach2013color,
  title={Color prediction with simplified Kubelka--Munk model in glazes containing Fe2O3--ZrSiO4 coral pink pigments},
  author={Schabbach, LM and Bondioli, Federica and Fredel, MC},
  journal={Dyes and pigments},
  volume={99},
  number={3},
  pages={1029--1035},
  year={2013},
  publisher={Elsevier}
}

@inproceedings{romagnoli2008neural,
  title={Neural network approach for color matching of ceramic glazes},
  author={Romagnoli, Marcello and Bondioli, Federica and Barattini, M and others},
  booktitle={International Congress of Ceramic Materiali},
  volume={1},
  pages={xx--xx},
  year={2008},
  organization={ECERS}
}

@article{vasic2025advanced,
  title={Advanced machine learning models for the prediction of ceramic tiles’ properties during the firing stage},
  author={Vasi{\'c}, Milica V and Awoyera, Paul O and Fadugba, Olaolu George and Bari{\v{s}}i{\'c}, Ivana and Grube{\v{s}}a, Ivanka Netinger},
  journal={Scientific reports},
  volume={15},
  number={1},
  pages={31397},
  year={2025},
  publisher={Nature Publishing Group UK London}
}

@misc{glazy,
title = {Glazy},
howpublished = {\url{https://glazy.org/}},
note = {Accessed: 2026-02-01},
year = {2026},
author = {{Glazy Contributors}},
url = {https://glazy.org/}
}

@article{ueki2018geochemical,
  title={Geochemical discrimination and characteristics of magmatic tectonic settings: A machine-learning-based approach},
  author={Ueki, Kenta and Hino, Hideitsu and Kuwatani, Tatsu},
  journal={Geochemistry, Geophysics, Geosystems},
  volume={19},
  number={4},
  pages={1327--1347},
  year={2018},
  publisher={Wiley Online Library}
}

@article{mues2025using,
  title={Using machine learning classifiers together with discrimination diagrams for validation of rock classification labels},
  author={Mues, Malte and Kraemer, Dennis and Styn, David M Ernst},
  journal={Applied Computing and Geosciences},
  pages={100288},
  year={2025},
  publisher={Elsevier}
}

@article{zhang2020data,
  title={Data-driven predictive models for chemical durability of oxide glass under different chemical conditions},
  author={Zhang, Yi and Li, Aize and Deng, Binghui and Hughes, Kelleen K},
  journal={npj Materials Degradation},
  volume={4},
  number={1},
  pages={14},
  year={2020},
  publisher={Nature Publishing Group UK London}
}

@article{ahmmad2021artificial,
  title={Artificial intelligence density model for oxide glasses},
  author={Ahmmad, Shaik Kareem and Jabeen, Nameera and Ahmed, Syed Taqi Uddin and Ahmed, Shaik Amer and Rahman, Syed},
  journal={Ceramics international},
  volume={47},
  number={6},
  pages={7946--7956},
  year={2021},
  publisher={Elsevier}
}

@article{trott2022random,
  title={Random forest rock type classification with integration of geochemical and photographic data},
  author={Trott, McLean and Leybourne, Matthew and Hall, Lindsay and Layton-Matthews, Daniel},
  journal={Applied Computing and Geosciences},
  volume={15},
  pages={100090},
  year={2022},
  publisher={Elsevier}
}

@article{ruiyi2021lithology,
  title={Lithology identification of igneous rocks based on XGboost and conventional logging curves, a case study of the eastern depression of Liaohe Basin},
  author={Ruiyi, HAN and Zhuwen, WANG and Wenhua, WANG and Fanghui, XU and Xinghua, QI and Yitong, CUI},
  journal={Journal of Applied Geophysics},
  volume={195},
  pages={104480},
  year={2021},
  publisher={Elsevier}
}

@article{xie2025prediction,
  title={Prediction of Thermal and Optical Properties of Oxyfluoride Glasses Based on Interpretable Machine Learning},
  author={Xie, Yuhao and Wang, Xiangfu},
  journal={Nanomaterials},
  volume={15},
  number={11},
  pages={860},
  year={2025},
  publisher={MDPI}
}

@article{feng2025application,
  title={Application of WOA optimized LightGBM in lithology identification of igneous logging},
  author={FENG, Huan and ZHANG, GuoQiang and CAO, Jun and REN, Hong and WAN, WenChun and LIU, DiRen},
  journal={Progress in Geophysics},
  volume={40},
  number={1},
  pages={230--242},
  year={2025},
  publisher={Progress in Geophysics}
}

@article{zhang2025geochemical,
  title={Geochemical Signatures and Element Interactions of Volcanic-Hosted Agates: Insights from Interpretable Machine Learning},
  author={Zhang, Peng and Xi, Xi and Wang, Bo-Chao},
  journal={Minerals},
  volume={15},
  number={9},
  pages={923},
  year={2025},
  publisher={MDPI}
}

@article{belciu2025ensemble,
  title={Ensemble Machine Learning for the Prediction and Understanding of the Refractive Index in Chalcogenide Glasses},
  author={Belciu, Miruna-Ioana and Velea, Alin},
  journal={Molecules},
  volume={30},
  number={8},
  pages={1745},
  year={2025},
  publisher={MDPI}
}

@article{bo2022real,
  title={Real-time hard-rock tunnel prediction model for rock mass classification using CatBoost integrated with Sequential Model-Based Optimization},
  author={Bo, Yin and Liu, Quansheng and Huang, Xing and Pan, Yucong},
  journal={Tunnelling and underground space technology},
  volume={124},
  pages={104448},
  year={2022},
  publisher={Elsevier}
}

@article{castela2010development,
  title={Development of coloured glazes for tile applications using Taguchi's method},
  author={Castela, AO and Fonseca, AT and Mantas, PQ},
  journal={Journal of the European Ceramic Society},
  volume={30},
  number={12},
  pages={2451--2455},
  year={2010},
  publisher={Elsevier}
}

@article{liu2021machine,
  title={Machine learning for glass science and engineering: A review},
  author={Liu, Han and Fu, Zipeng and Yang, Kai and Xu, Xinyi and Bauchy, Mathieu},
  journal={Journal of Non-Crystalline Solids},
  volume={557},
  pages={119419},
  year={2021},
  publisher={Elsevier}
}

@article{feng2023phase,
  title={Phase-separated Tenmoku “Blue” glaze: Microstructure and coloring mechanism},
  author={Feng, Li and Wang, Fen and Luo, Hongjie and Zhu, Jianfeng and Wang, Minli and Yang, Chi and Sun, Jianxing and Wang, Tian},
  journal={Journal of the European Ceramic Society},
  volume={43},
  number={14},
  pages={6581--6589},
  year={2023},
  publisher={Elsevier}
}

@article{wu2021firing,
  title={Firing process and colouring mechanism of black glaze and brown glaze porcelains from the Yuan and Ming dynasties from the Qingliang Temple kiln in Baofeng, Henan, China},
  author={Wu, Bo and Zhao, Weijuan and Ren, Xiao and Liu, Xiaomin and Li, Bo and Feng, Songlin and Feng, Xiangqian and Zhao, Hong},
  journal={Ceramics International},
  volume={47},
  number={23},
  pages={32817--32827},
  year={2021},
  publisher={Elsevier}
}

@article{zhao2023revealing,
  title={Revealing the individual effects of firing temperature and chemical composition on raman parameters of celadon glaze},
  author={Zhao, Lan and Zhang, Yunjun},
  journal={Ceramics},
  volume={6},
  number={2},
  pages={1263--1276},
  year={2023},
  publisher={MDPI}
}

@article{santos2025temperature,
  title={Temperature Assessment Through Decal Color in Microwave-Fired Porcelain},
  author={Santos, Tiago and Hennetier, Luc and Costa, V{\'\i}tor AF and Costa, Lu{\'\i}s C},
  journal={Journal of Manufacturing and Materials Processing},
  volume={9},
  number={7},
  pages={213},
  year={2025},
  publisher={MDPI}
}

@article{rumble2017accessing,
  title={Accessing materials data: challenges and directions in the digital era},
  author={Rumble Jr, John R},
  journal={Integrating Materials and Manufacturing Innovation},
  volume={6},
  number={2},
  pages={172--186},
  year={2017},
  publisher={Springer}
}

@article{fujinuma2022big,
  title={Why big data and compute are not necessarily the path to big materials science},
  author={Fujinuma, Naohiro and DeCost, Brian and Hattrick-Simpers, Jason and Lofland, Samuel E},
  journal={Communications Materials},
  volume={3},
  number={1},
  pages={59},
  year={2022},
  publisher={Nature Publishing Group UK London}
}

@article{chakraborty2026survey,
  title={A survey of AI-supported materials informatics},
  author={Chakraborty, Sanjay and Bj{\"o}rk, Jonas and Dahlqvist, Martin and Rosen, Johanna and Heintz, Fredrik},
  journal={Computer Science Review},
  volume={59},
  pages={100845},
  year={2026},
  publisher={Elsevier}
}

@article{imer2016effects,
  title={Effects of firing temperatures and compositions on the formation of nano particles in lustre layers on a lead-alkali glaze},
  author={Imer, C and G{\"u}nay, E and {\"O}ve{\c{c}}o{\u{g}}lu, ML},
  journal={Ceramics International},
  volume={42},
  number={15},
  pages={17222--17228},
  year={2016},
  publisher={Elsevier}
}

@article{rother2004grabcut,
  title={" GrabCut" interactive foreground extraction using iterated graph cuts},
  author={Rother, Carsten and Kolmogorov, Vladimir and Blake, Andrew},
  journal={ACM transactions on graphics (TOG)},
  volume={23},
  number={3},
  pages={309--314},
  year={2004},
  publisher={ACM New York, NY, USA}
}

@article{fu2025digital,
  title={Digital color enhancement in ceramic imagery using graph-guided residual learning and adaptive scattering models},
  author={Fu, Zhi},
  journal={Journal of Computational Methods in Sciences and Engineering},
  pages={14727978251391297},
  year={2025},
  publisher={SAGE Publications Sage UK: London, England}
}

@article{wei2020experimental,
  title={Experimental study on glaze icing detection of 110 kV composite insulators using fiber Bragg gratings},
  author={Wei, Jie and Hao, Yanpeng and Fu, Yuan and Yang, Lin and Gan, Jiulin and Li, Han},
  journal={Sensors},
  volume={20},
  number={7},
  pages={1834},
  year={2020},
  publisher={MDPI}
}

@article{ojala2002multiresolution,
  title={Multiresolution gray-scale and rotation invariant texture classification with local binary patterns},
  author={Ojala, Timo and Pietikainen, Matti and Maenpaa, Topi},
  journal={IEEE Transactions on pattern analysis and machine intelligence},
  volume={24},
  number={7},
  pages={971--987},
  year={2002},
  publisher={IEEE}
}

@inproceedings{gao2010improved,
  title={An improved Sobel edge detection},
  author={Gao, Wenshuo and Zhang, Xiaoguang and Yang, Lei and Liu, Huizhong},
  booktitle={2010 3rd International conference on computer science and information technology},
  volume={5},
  pages={67--71},
  year={2010},
  organization={IEEE}
}

@inproceedings{yan2016attribute2image,
  title={Attribute2image: Conditional image generation from visual attributes},
  author={Yan, Xinchen and Yang, Jimei and Sohn, Kihyuk and Lee, Honglak},
  booktitle={European conference on computer vision},
  pages={776--791},
  year={2016},
  organization={Springer}
}

@article{yamagiwa2025analytical,
  title={An Analytical Model using CVAE-based Image Generation from Product Descriptions and Image Data},
  author={Yamagiwa, Ayako and Goto, Masayuki and others},
  journal={Industrial Engineering \& Management Systems},
  volume={24},
  number={4},
  pages={650--662},
  year={2025}
}

@article{wang2025lightweight,
  title={Lightweight Text-to-Image Generation Model Based on Contrastive Language-Image Pre-Training Embeddings and Conditional Variational Autoencoders},
  author={Wang, Yubo and Zhang, Gaofeng},
  journal={Electronics},
  volume={14},
  number={11},
  pages={2185},
  year={2025},
  publisher={MDPI}
}

@inproceedings{ding2021ccgan,
  title={Ccgan: Continuous conditional generative adversarial networks for image generation},
  author={Ding, Xin and Wang, Yongwei and Xu, Zuheng and Welch, William J and Wang, Z Jane},
  booktitle={International conference on learning representations},
  year={2021}
}

@inproceedings{jin2019image,
  title={Image generation method based on improved condition GAN},
  author={Jin, Qiuzi and Luo, Xin and Shi, Youqun and Kita, Kenji},
  booktitle={2019 6th international conference on systems and informatics (ICSAI)},
  pages={1290--1294},
  year={2019},
  organization={IEEE}
}

@article{li2020sar,
  title={A SAR-to-optical image translation method based on conditional generation adversarial network (cGAN)},
  author={Li, Yu and Fu, Randi and Meng, Xiangchao and Jin, Wei and Shao, Feng},
  journal={Ieee Access},
  volume={8},
  pages={60338--60343},
  year={2020},
  publisher={IEEE}
}

@article{riedel2022automated,
  title={Automated quality control of vacuum insulated glazing by convolutional neural network image classification},
  author={Riedel, Henrik and Mokdad, Sleheddine and Schulz, Isabell and Kocer, Cenk and Rosendahl, Philipp L and Schneider, Jens and Kraus, Michael A and Drass, Michael},
  journal={Automation in Construction},
  volume={135},
  pages={104144},
  year={2022},
  publisher={Elsevier}
}

@article{mao2025quantitative,
  title={A quantitative study of phase assemblage in cement-fly ash-slag ternary systems using machine learning-assisted BSE-EDS image analysis},
  author={Mao, Li-xuan and He, Fuqiang and Li, Lihui and Xu, Wei and Wang, Yong and Liu, Qing-feng},
  journal={Construction and Building Materials},
  volume={498},
  pages={143712},
  year={2025},
  publisher={Elsevier}
}

@article{zhang2021understanding,
  title={Understanding the role of glaze layer with aligned images from multiple surface characterization techniques},
  author={Zhang, Chuchu and Neu, Richard W},
  journal={Wear},
  volume={477},
  pages={203837},
  year={2021},
  publisher={Elsevier}
}

@article{chen2020cantonese,
  title={Cantonese porcelain image generation using user-guided generative adversarial networks},
  author={Chen, Steven Szu-Chi and Cui, Hui and Tan, Peng and Sun, Xiaohong and Ji, Yi and Duh, Henry},
  journal={IEEE Computer Graphics and Applications},
  volume={40},
  number={5},
  pages={100--107},
  year={2020},
  publisher={IEEE}
}

\newpage
\appendix
\section{Appendix A: Data Preprocessing Details}

\subsection{Color Annotation Methodology}
Transparency and surface-texture labels are obtained directly from structured dropdown menus on the Glazy website and therefore require no additional post-validation. In contrast, the dataset contains two automatically recognized color-related fields whose accuracy cannot be reliably quantified. We therefore apply additional processing to construct a higher-confidence color annotation set for both training and evaluation.

\paragraph{Test set (manual curation).}
We first collected 8{,}000 candidate samples. For each sample, we manually selected the most representative glaze photograph (when multiple images were available) and assigned the corresponding glaze color. This process produced 4{,}903 manually labeled test samples. The remaining 3{,}097 samples were judged as \emph{uncertain} (ambiguous color) or \emph{poor quality} (e.g., lighting/coverage issues) and were excluded from the test set; instead, they were moved to the training pool to assess whether our training-set filtering procedure could remove them automatically.

\paragraph{Training set (model-assisted filtering).}
Using the manually labeled subset as a reference, we filter color annotations in the training set as follows:

\begin{enumerate}
    \item \textbf{Reference model (ensemble construction).}
    We train and compare four machine-learning models to learn the recipe-to-color mapping from the manually labeled data. The two best-performing models---Random Forest and XGBoost---are retained and combined into an ensemble for downstream color selection.

    \item \textbf{RGB-based agreement and selection.}
    The two models independently predict an RGB color. Let the two predicted candidates be \(\mathbf{c}_1,\mathbf{c}_2\in\mathbb{R}^3\), and let \(\bar{\mathbf{c}}_{\mathrm{pred}}\) denote their centroid. We compute Euclidean distances
    \[
        d_k=\bigl\|\mathbf{c}_k-\bar{\mathbf{c}}_{\mathrm{pred}}\bigr\|_2,\quad k\in\{1,2\},
        \qquad \text{and select } \arg\min_k d_k.
    \]
    Intuitively, this step prefers the candidate closer to the consensus of the two predictors.

    \item \textbf{Ambiguity filtering.}
    If \(|d_1-d_2|<10\), the two candidates are considered equally plausible and the sample is marked as ambiguous and discarded. After filtering, 12{,}175 training samples remain with validated color annotations.
\end{enumerate}

\paragraph{Sanity check.}
All 3{,}097 samples previously marked as \emph{uncertain} during manual curation are removed by the above filtering pipeline, supporting the effectiveness of the ambiguity criteria.

\subsection{Feature Representation}
The two benchmark categories---\emph{property prediction} and \emph{image generation}---operate on different input spaces. Accordingly, we use task-specific feature representations.

\subsubsection{Property Prediction Tasks (Tasks A, B, C1, C2): Recipe Feature Vector}
All property prediction models take a 22-dimensional recipe feature vector \(\mathbf{x}_i\in\mathbb{R}^{22}\) as input, encoding glaze chemistry and firing conditions. The vector consists of three parts:

\paragraph{UMF oxides (18 dimensions).}
Recipes are represented in UMF notation (flux group normalized to unity). We track 18 oxides:
SiO\(_2\), Al\(_2\)O\(_3\), B\(_2\)O\(_3\), Li\(_2\)O, Na\(_2\)O, K\(_2\)O, MgO, CaO, SrO, BaO, ZnO, TiO\(_2\),
Fe\(_2\)O\(_3\), P\(_2\)O\(_5\), SnO\(_2\), Cr\(_2\)O\(_3\), ZrO\(_2\), and PbO.
This oxide set is derived from the UMF keys observed in the dataset. Missing oxides are represented as zeros.

\paragraph{Cone range (2 dimensions).}
The firing temperature range is encoded as \texttt{cone\_min} and \texttt{cone\_max}. A single cone specification sets the two values equal.

\paragraph{Atmosphere (2 dimensions).}
Atmosphere is encoded with two binary indicators: Oxidation and Reduction. Missing or non-standard atmosphere entries are encoded as all zeros.

\subsubsection{Image Generation Task (Task D): Visual Condition Vector}
For Task D, the conditional generative models (cVAE and cGAN) do not use recipe chemistry. Instead, they condition image synthesis on a 25-dimensional visual attribute vector \(\mathbf{c}_i\in\mathbb{R}^{25}\), formed by concatenating four components (Table~\ref{tab:condition_vector}).

\begin{table}[htbp]
\centering
\caption{Composition of the 25-dimensional condition vector for Task D}
\label{tab:condition_vector}
\begin{tabular}{llcc}
\toprule
\textbf{Component} & \textbf{Encoding} & \textbf{Dim.} & \textbf{Value range} \\
\midrule
Surface texture & One-hot (9 classes) & 9 & \(\{0,1\}^9\) \\
Transparency & One-hot (4 classes) & 4 & \(\{0,1\}^4\) \\
Color family & One-hot (9 classes) & 9 & \(\{0,1\}^9\) \\
RGB color & Normalized continuous & 3 & \([0,1]^3\) \\
\midrule
\textbf{Total} &  & \textbf{25} &  \\
\bottomrule
\end{tabular}
\end{table}

RGB values are normalized by dividing by 255. Samples used for image-generation training must have all four attributes available; consequently, the effective training set corresponds to the intersection of surface, transparency, color-family, and RGB annotations.

\newpage
\section{Appendix B: Task Statistics}

\subsection{Transparency Category Distribution}

The transparency attribute partitions samples into four mutually exclusive categories: Opaque, Semi-opaque, Translucent, and Transparent. This attribute is typically associated with a material’s degree of light transmission and scattering, and is therefore a key factor in appearance understanding tasks. \cref{tab:transparency_dist} reports the transparency-category distribution in the train and test sets. We can observe that: (1) \textbf{Opaque} is the dominant category, accounting for \(48.6\%\) of the training set and \(43.3\%\) of the test set, indicating that the dataset is overall biased toward opaque materials; (2) the remaining three categories constitute \(21.8\%\), \(13.3\%\), and \(16.3\%\) of the training set, and \(22.8\%\), \(17.9\%\), and \(15.9\%\) of the test set, respectively, reflecting a moderately imbalanced distribution; and (3) while the overall distributional patterns are largely consistent between training and testing, the test set contains a relatively higher proportion of \textbf{Translucent} samples (\(17.9\%\) vs. \(13.3\%\)), suggesting that evaluation involves more translucent cases and thus places greater demands on model generalization.

\begin{table}[h]
\centering
\caption{Transparency category distribution}
\label{tab:transparency_dist}
\begin{tabular}{lrrrr}
\toprule
\textbf{Category} & \textbf{Train Count} & \textbf{Train Ratio} & \textbf{Test Count} & \textbf{Test Ratio} \\
\midrule
Opaque & 3,200 & 48.6\% & 1,440 & 43.3\% \\
Semi-opaque & 1,438 & 21.8\% & 759 & 22.8\% \\
Translucent & 874 & 13.3\% & 595 & 17.9\% \\
Transparent & 1,072 & 16.3\% & 528 & 15.9\% \\
\midrule
\textbf{Total} & 6,584 & 100.0\% & 3,322 & 100.0\% \\
\bottomrule
\end{tabular}
\end{table}

\subsection{Surface Category Distribution}

The surface attribute characterizes material surfaces in terms of specular versus diffuse reflectance under illumination, as well as differences in microscopic roughness. It comprises nine categories, including Glossy, Semi-glossy, Matte, and Satin, etc. This dimension is more fine-grained, with category boundaries closer to a perceptual continuum; consequently, it is often more sensitive to data coverage and annotation consistency. \cref{tab:surface_dist} reports the distribution of surface categories in the train and test sets. We observe that: (1) \textbf{Glossy} is the dominant category, accounting for nearly half of the training set (\(49.1\%\)) and remaining the most prevalent in the test set (\(45.4\%\)), indicating a bias toward specular or highly reflective surfaces; (2) mid-frequency categories (e.g., Semi-glossy, Matte, and Satin) each fall in the \(8\%\sim 15\%\) range, providing a reasonable degree of diversity for learning; and (3) tail categories (e.g., Dry Matte and Stony Matte) each comprise less than \(2\%\), exhibiting a \textbf{long-tailed} distribution that may lead to lower recall or unstable decision boundaries for these classes. The training and test splits follow similar trends for the major categories, while several fine-grained classes appear slightly more frequent in the test set (e.g., Semi-glossy, Matte, and Satin-matte), suggesting that evaluation difficulty is not determined solely by head classes.

\begin{table}[h]
\centering
\caption{Surface category distribution}
\label{tab:surface_dist}
\begin{tabular}{lrrrr}
\toprule
\textbf{Category} & \textbf{Train Count} & \textbf{Train Ratio} & \textbf{Test Count} & \textbf{Test Ratio} \\
\midrule
Glossy & 3,360 & 49.1\% & 1,692 & 45.4\% \\
Semi-glossy & 856 & 12.5\% & 539 & 14.5\% \\
Matte & 732 & 10.7\% & 462 & 12.4\% \\
Satin & 623 & 9.1\% & 324 & 8.7\% \\
Satin-matte & 394 & 5.8\% & 284 & 7.6\% \\
Semi-matte & 382 & 5.6\% & 223 & 6.0\% \\
Smooth Matte & 239 & 3.5\% & 122 & 3.3\% \\
Dry Matte & 137 & 2.0\% & 33 & 0.9\% \\
Stony Matte & 121 & 1.8\% & 51 & 1.4\% \\
\midrule
\textbf{Total} & 6,844 & 100.0\% & 3,730 & 100.0\% \\
\bottomrule
\end{tabular}
\end{table}

\subsection{Color Family Category Distribution}

The color-family attribute captures the dominant chromatic family of a material or object, covering nine categories such as Orange, Gray, and Blue, etc. Since color is strongly coupled with imaging conditions, including illumination, white balance, and background interference. Its statistics are useful for diagnosing potential color bias in data collection/annotation and for identifying distribution shifts between training and testing. \cref{tab:color_family_dist} summarizes the counts and ratios of color-family categories in the training and test sets. We can observe clear category bias: (1) \textbf{Orange} and \textbf{Gray} are over-represented in the training set (\(36.0\%\) and \(20.3\%\), respectively), indicating that the data are more concentrated in warm and neutral tones; (2) in the test set, the proportion of \textbf{Orange} drops to \(27.1\%\), whereas \textbf{Blue} increases to \(20.2\%\) (vs. \(16.1\%\) in training). Meanwhile, categories such as \textbf{White/Black/Red/Green} also rise to varying degrees, reflecting a noticeable train-test distribution shift; and (3) \textbf{Purple} is rare in both splits (below \(1\%\)), constituting a typical tail class. These characteristics suggest that if a model over-relies on color priors during training, its generalization may degrade when the test distribution changes. Therefore, in the experiments, it is advisable to incorporate class-balancing strategies, color-based augmentation, or stratified metrics to more comprehensively assess model performance.

\begin{table}[h]
\centering
\caption{Color Family category distribution}
\label{tab:color_family_dist}
\begin{tabular}{lrrrr}
\toprule
\textbf{Category} & \textbf{Train Count} & \textbf{Train Ratio} & \textbf{Test Count} & \textbf{Test Ratio} \\
\midrule
Orange & 4,377 & 36.0\% & 1,327 & 27.1\% \\
Gray & 2,469 & 20.3\% & 775 & 15.8\% \\
Blue & 1,959 & 16.1\% & 991 & 20.2\% \\
Yellow & 1,203 & 9.9\% & 460 & 9.4\% \\
Green & 789 & 6.5\% & 380 & 7.8\% \\
Red & 668 & 5.5\% & 371 & 7.6\% \\
White & 501 & 4.1\% & 357 & 7.3\% \\
Black & 130 & 1.1\% & 201 & 4.1\% \\
Purple & 79 & 0.6\% & 41 & 0.8\% \\
\midrule
\textbf{Total} & 12,175 & 100.0\% & 4,903 & 100.0\% \\
\bottomrule
\end{tabular}
\end{table}

\newpage
\section{Appendix C: Implementation Details of LLM Baselines}
This appendix details the implementation of the LLM baselines used for three classification tasks: \emph{transparency}, \emph{surface texture}, and \emph{color family}. We describe the API setup, input serialization, prompt design (zero-shot and few-shot), few-shot example selection, and response parsing.

\subsection{Model and API Configuration}
All experiments are executed through the OpenRouter API using an OpenAI-compatible SDK. We evaluate three models: \texttt{GPT-4o-mini}, \texttt{DeepSeek-v3}, and \texttt{Claude Sonnet 4.5}. Unless otherwise stated, all models share the same inference and runtime settings:

\begin{itemize}
    \item \textbf{Decoding:} temperature \(= 0.0\) (deterministic decoding), \texttt{max\_tokens} \(= 500\).
    \item \textbf{Reliability:} per-request timeout \(= 60\) seconds; up to 3 retries with exponential backoff (waiting 1, 2, and 4 seconds).
    \item \textbf{Throughput:} for each task, predictions over all test samples are issued concurrently using \texttt{ThreadPoolExecutor} with up to 20 worker threads.
    \item \textbf{Checkpointing:} intermediate outputs are saved every 10 predictions to enable resumable execution.
\end{itemize}

All models are used \emph{as-is} (no fine-tuning, no additional training, and no task-specific adaptation).

\subsection{Input Representation}
Each test sample is serialized into three textual blocks and injected into the prompt:

\begin{enumerate}
    \item \textbf{Chemical composition (wt.\% oxides).}
    All oxide weight percentages larger than \(0.01\%\) are listed in the format
    \texttt{Oxide: value\%} (comma-separated), e.g.,
    \texttt{SiO2: 45.20\%, Al2O3: 12.80\%, CaO: 8.50\%, \ldots}.
    \item \textbf{UMF formula.}
    All UMF entries larger than \(0.01\) are listed as \texttt{Oxide: value} and prefixed by \texttt{UMF Formula:}.
    \item \textbf{Firing parameters.}
    If available, we include cone information (\texttt{Cone: N} or \texttt{Cone Range: N--M}) and atmosphere (\texttt{Oxidation} or \texttt{Reduction}).
    Otherwise, the field is set to \texttt{No additional firing parameters available.}
\end{enumerate}

\subsection{Prompt Design}
For each task, we use a unified prompt template that supports both zero-shot and few-shot evaluation. The template consists of:

\begin{enumerate}
    \item a role declaration and task instruction;
    \item an explicit, enumerated label set with short descriptions;
    \item domain rules connecting oxides/firing conditions to visual properties;
    \item an optional few-shot block \texttt{\{few\_shot\_examples\}};
    \item the query sample (three input blocks as above);
    \item a strict output constraint: \emph{output exactly one label from the allowed set}.
\end{enumerate}

For zero-shot evaluation (\(K=0\)), the few-shot block is omitted. For \(K\)-shot evaluation, the block is populated as described in Section~\ref{sec:fewshot_selection}.

\paragraph{Task-specific instantiations.}
The three tasks share the same structure but differ in label sets and domain rules:

\begin{itemize}
    \item \textbf{Transparency (4 classes).}
    Labels: \emph{Transparent}, \emph{Translucent}, \emph{Semi-opaque}, \emph{Opaque}.
    Rules emphasize that higher SiO\(_2\) tends to increase transparency, while TiO\(_2\)/SnO\(_2\)/ZrO\(_2\) are commonly associated with opacity.
    \item \textbf{Surface texture (9 classes).}
    Labels: \emph{Glossy}, \emph{Semi-glossy}, \emph{Satin}, \emph{Satin-matte}, \emph{Matte}, \emph{Semi-matte}, \emph{Smooth Matte}, \emph{Dry Matte}, \emph{Stony Matte}.
    Rules relate higher SiO\(_2\) relative to fluxes to glossier surfaces, higher Al\(_2\)O\(_3\)/MgO to matte characteristics, and ZnO to satin-like surfaces.
    \item \textbf{Color family (9 classes).}
    Labels: \emph{Black}, \emph{Blue}, \emph{Gray}, \emph{Green}, \emph{Orange}, \emph{Purple}, \emph{Red}, \emph{White}, \emph{Yellow}.
    Rules map typical colorants to hues, e.g., Fe\(_2\)O\(_3\) \(\rightarrow\) red/brown (oxidation) or blue/gray (reduction); CoO \(\rightarrow\) blue; CuO \(\rightarrow\) green (oxidation) or red (reduction); MnO \(\rightarrow\) purple; Cr\(_2\)O\(_3\) \(\rightarrow\) green; TiO\(_2\) \(\rightarrow\) white.
\end{itemize}

\subsection{Few-shot Example Selection}
\label{sec:fewshot_selection}
For \(K\)-shot evaluation, we set \(K=5\). Few-shot examples are selected from the training set using a stratified round-robin strategy:

\begin{enumerate}
    \item Collect training samples that (i) have valid labels for the target task and (ii) contain non-empty chemical composition data. Group them by class.
    \item Iterate classes in insertion order and draw one example per class in sequence until \(K\) examples are obtained. Classes with no remaining samples are removed from the rotation.
    \item Serialize each selected example using the same three-block format as the query, followed by \texttt{Answer: \{label\}}.
\end{enumerate}

This procedure encourages class coverage in-context, ensuring up to \(\min(K, |\mathcal{C}|)\) distinct classes appear in the prompt. This is particularly relevant for imbalanced tasks (e.g., surface texture, where \emph{Glossy} accounts for 49\% of samples).

\paragraph{Few-shot block format.}
Each example follows the structure below:
\begin{verbatim}
Example i:
Chemical Composition: SiO2: 45.20%, Al2O3: 12.80%, CaO: 8.50%, ...
UMF Formula: SiO2: 3.21, Al2O3: 0.43, CaO: 0.55, ...
Firing Cone: 6
Firing Atmosphere: Oxidation
Answer: Translucent
\end{verbatim}

\subsection{Response Parsing}
We parse model outputs using case-insensitive label matching on the first line of the response:

\begin{enumerate}
    \item Strip leading/trailing whitespace and quotation characters, then extract the first line.
    \item Iterate through the ordered list of valid labels and return the first label whose lowercase form appears as a substring of the lowercase response line.
    \item For multi-word labels (e.g., \emph{Semi-opaque}, \emph{Satin-matte}, \emph{Smooth Matte}), we accept both hyphenated and space-separated variants.
    \item Outputs that match none of the valid labels are recorded as parsing failures and excluded from metric computation.
\end{enumerate}

\newpage
\section{Appendix D: Specifications of Image-Generation Baselines}
This appendix reports the technical specifications of two baseline models for the conditional glaze image generation task (Task D), including the problem formulation, model architectures, training objectives, hyperparameters, and data preprocessing.

\subsection{Problem Formulation}
We learn a conditional generative model of the form \(p_{\theta}(\mathbf{x}\mid \mathbf{c})\), where
\(\mathbf{x}\in\mathbb{R}^{3\times128\times128}\) denotes the generated glaze surface image and
\(\mathbf{c}\in\mathbb{R}^{25}\) denotes the condition vector encoding visual attributes only. Specifically,
\[
\mathbf{c}
=
\Bigl[
\underbrace{\mathbf{s}}_{\text{surface, }9},\;
\underbrace{\mathbf{t}}_{\text{transparency, }4},\;
\underbrace{\mathbf{f}}_{\text{color family, }9},\;
\underbrace{\tilde{\mathbf{r}}}_{\text{RGB, }3}
\Bigr]\in\mathbb{R}^{25},
\]
where \(\mathbf{s}\), \(\mathbf{t}\), and \(\mathbf{f}\) are one-hot vectors, and
\(\tilde{\mathbf{r}}=\mathbf{r}/255\in[0,1]^3\) is the normalized RGB value. Firing parameters (cone and atmosphere) are not used in Task D, since generation is conditioned solely on the target visual appearance.

\subsection{Baseline Models}
We evaluate two parametric conditional generative baselines: a conditional variational autoencoder (cVAE) and a lightweight conditional GAN (cGAN).

\subsubsection{Conditional Variational Autoencoder (cVAE)}
\paragraph{Architecture.}
The encoder maps an image--condition pair \((\mathbf{x},\mathbf{c})\) to a Gaussian posterior
\(q_{\phi}(\mathbf{z}\mid \mathbf{x},\mathbf{c})\).
The image branch applies four strided convolutional blocks
(\(3\to32\to64\to128\to256\) channels; kernel \(4\times4\), stride 2, padding 1; each followed by BatchNorm and ReLU),
yielding a \(256\times 8\times 8\) feature map. The feature map is flattened to 16{,}384 dimensions.
The condition vector \(\mathbf{c}\) is embedded by an MLP layer (\(25\to128\), ReLU). The concatenated representation
(\(16{,}384+128=16{,}512\) dims) is mapped to \(\boldsymbol{\mu}\in\mathbb{R}^{64}\) and
\(\log\boldsymbol{\sigma}^2\in\mathbb{R}^{64}\) via two independent linear heads.

The decoder takes \([\mathbf{z};\mathbf{c}]\in\mathbb{R}^{89}\), projects it to \(256\times 8\times 8\) through a linear layer,
and then applies four transposed convolutional blocks
(\(256\to128\to64\to32\to3\), same kernel/stride/padding as above). A final Tanh activation produces outputs in \([-1,1]\).

\paragraph{Objective.}
Training maximizes the evidence lower bound (ELBO), implemented as an MSE reconstruction term plus a weighted KL regularizer:
\[
\mathcal{L}_{\mathrm{cVAE}}
=
\underbrace{\frac{1}{N}\sum_{i=1}^{N}\|\mathbf{x}_i-\hat{\mathbf{x}}_i\|^2}_{\text{reconstruction (MSE)}}
-
\beta\cdot
\underbrace{\frac{1}{N}\sum_{i=1}^{N}\sum_{j=1}^{64}
\bigl(1+\log\sigma_{ij}^2-\mu_{ij}^2-\sigma_{ij}^2\bigr)}_{\mathrm{KL}\;\text{to }p(\mathbf{z})=\mathcal{N}(\mathbf{0},\mathbf{I})}.
\]
We set \(\beta=10^{-4}\) to mitigate posterior collapse under limited training data.

\paragraph{Hyperparameters.}
Table~\ref{tab:vae_hyper} summarizes the cVAE configuration.

\begin{table}[htbp]
\centering
\caption{Hyperparameter configuration for cVAE}
\label{tab:vae_hyper}
\begin{tabular}{ll}
\toprule
\textbf{Parameter} & \textbf{Value} \\
\midrule
Latent dimension \(d_z\) & 64 \\
Condition dimension \(d_c\) & 25 \\
KL weight \(\beta\) & \(10^{-4}\) \\
Optimizer & Adam \\
Learning rate & \(5\times 10^{-4}\) \\
LR scheduler & ReduceLROnPlateau (factor 0.5, patience 10) \\
Batch size & 32 \\
Training epochs & 200 \\
\bottomrule
\end{tabular}
\end{table}

\subsubsection{Lightweight Conditional GAN (cGAN)}
\paragraph{Generator.}
The generator \(G:\mathbb{R}^{64}\times\mathbb{R}^{25}\to\mathbb{R}^{3\times128\times128}\) concatenates
noise \(\mathbf{z}\sim\mathcal{N}(\mathbf{0},\mathbf{I})\) and condition \(\mathbf{c}\) into an 89-dimensional vector.
It is projected to \(128\times 8\times 8\) via a linear layer and upsampled through four transposed convolutional blocks
(\(128\to64\to32\to16\to3\); kernel \(4\times4\), stride 2, padding 1) with BatchNorm and ReLU activations.
The output layer uses Tanh to match the \([-1,1]\) normalized image range.

\paragraph{Discriminator (critic).}
The discriminator \(D:\mathbb{R}^{3\times128\times128}\times\mathbb{R}^{25}\to\mathbb{R}\) injects conditioning by
expanding \(\mathbf{c}\) to a spatial map: a linear layer (\(25\to 128\times128\)) reshaped to
\(1\times128\times128\), concatenated with the input image to form a \(4\times128\times128\) tensor.
The critic then applies four strided convolutions
(\(4\to16\to32\to64\to128\) channels; kernel \(4\times4\), stride 2) with LeakyReLU(0.2),
reducing to an \(8\times 8\) feature map. A final \(8\times 8\) convolution produces a scalar score.
We omit BatchNorm in the critic to improve stability under small batch sizes.

\paragraph{Objective (WGAN-GP).}
We adopt WGAN-GP. The critic minimizes
\[
\mathcal{L}_D
=
-\mathbb{E}_{\mathbf{x}}[D(\mathbf{x},\mathbf{c})]
+\mathbb{E}_{\mathbf{z}}[D(G(\mathbf{z},\mathbf{c}),\mathbf{c})]
+\lambda_{\mathrm{gp}}\,
\mathbb{E}_{\hat{\mathbf{x}}}\Bigl[\bigl(\|\nabla_{\hat{\mathbf{x}}}D(\hat{\mathbf{x}},\mathbf{c})\|_2-1\bigr)^2\Bigr],
\]
where \(\hat{\mathbf{x}}=\epsilon\,\mathbf{x}_{\mathrm{real}}+(1-\epsilon)\,\mathbf{x}_{\mathrm{fake}}\),
\(\epsilon\sim\mathrm{Uniform}(0,1)\), and \(\lambda_{\mathrm{gp}}=10\).
The generator minimizes
\(\mathcal{L}_G=-\mathbb{E}_{\mathbf{z}}[D(G(\mathbf{z},\mathbf{c}),\mathbf{c})]\).

\paragraph{Hyperparameters.}
Table~\ref{tab:gan_hyper} lists the cGAN configuration.

\begin{table}[htbp]
\centering
\caption{Hyperparameter configuration for cGAN}
\label{tab:gan_hyper}
\begin{tabular}{ll}
\toprule
\textbf{Parameter} & \textbf{Value} \\
\midrule
Noise dimension \(d_z\) & 64 \\
Condition dimension \(d_c\) & 25 \\
Gradient penalty \(\lambda_{\mathrm{gp}}\) & 10 \\
Critic steps per generator step \(n_{\mathrm{critic}}\) & 3 \\
Optimizer & Adam (\(\beta_1=0.5\), \(\beta_2=0.999\)) \\
Learning rate (G and D) & \(2\times 10^{-4}\) \\
Batch size & 32 \\
Training epochs & 250 \\
\bottomrule
\end{tabular}
\end{table}

\subsection{Training Data and Preprocessing}
The image generation subset includes samples with complete annotations for all four conditioning components
(surface, transparency, color family, and RGB). We use a 90/10 random split within the training portion,
resulting in approximately 3{,}640 training samples and 400 validation samples; the test set contains 438 held-out samples.

Images are preprocessed as follows:
\begin{enumerate}
    \item Resize to \(128\times128\) using Lanczos resampling.
    \item Normalize pixel values to \([-1,1]\) via \((x/255-0.5)/0.5\).
    \item Apply random horizontal flipping (probability 0.5) to training images only.
\end{enumerate}

\end{document}